# An Improved Admissible Heuristic for Learning Optimal Bayesian Networks

**Changhe Yuan**[1,3] **and Brandon Malone**[2,3]
[1]Queens College/City University of New York, [2]Helsinki Institute for Information Technology,
[3]and Mississippi State University
changhe.yuan@qc.cuny.edu, bmmalone@gmail.com

## Abstract

Recently two search algorithms, A* and breadth-first branch and bound (BFBnB), were developed based on a simple admissible heuristic for learning Bayesian network structures that optimize a scoring function. The heuristic represents a relaxation of the learning problem such that each variable chooses optimal parents independently. As a result, the heuristic may contain many directed cycles and result in a loose bound. This paper introduces an improved admissible heuristic that tries to avoid directed cycles within small groups of variables. A sparse representation is also introduced to store only the unique optimal parent choices. Empirical results show that the new techniques significantly improved the efficiency and scalability of A* and BFBnB on most of datasets tested in this paper.

## 1 Introduction

Bayesian networks are often used to represent relationships among variables in a domain. When the network structure is unknown, we can learn the structure directly from a given dataset. Several exact algorithms for learning optimal Bayesian networks have been developed based on dynamic programming (Koivisto and Sood 2004; Ott, Imoto, and Miyano 2004; Silander and Myllymaki 2006; Singh and Moore 2005; Parviainen and Koivisto 2009; Malone, Yuan, and Hansen 2011), branch and bound (de Campos and Ji 2011), and linear and integer programming (Cussens 2011; Jaakkola et al. 2010).

Recently, Yuan *et al.* (2011) proposed a shortest-path finding formulation for the Bayesian network structure learning problem, in which the shortest path found in an implicit search graph called order graph corresponds to an optimal network structure. An A* search algorithm was developed to solve the search problem. Malone *et al.* (2011) adopted the same formulation, but realized that the search can be performed in a layered, breadth-first order. External memory and delayed duplicate detection are used to ensure completion regardless of the amount of available RAM. This algorithm, named breadth-first branch and bound (BFBnB), was shown to have similar runtimes as A* but scale to many more variables.

A simple admissible heuristic was used in the A* and BFBnB algorithms (Yuan, Malone, and Wu 2011; Malone et al. 2011) to guide the search. Its main idea is to relax the acyclicity constraint of Bayesian networks such that each variable can freely choose optimal parents from all the other variables. The heuristic provides an optimistic estimation on how good a solution can be and, hence, is admissible. However, the simple relaxation behind the heuristic may introduce many directed cycles and result in a loose bound.

This paper introduces a much improved admissible heuristic named *k-cycle conflict heuristic* based on the additive pattern database technique (Felner, Korf, and Hanan 2004). The main idea is to avoid directed cycles within small groups of variables, called patterns, and compute heuristic values by concatenating patterns. Also, a set of exponential-size parent graphs were created by the A* and BFBnB algorithms to retrieve optimal parent choices during the search. We introduce a sparse representation for storing only unique optimal parent sets which can improve both time and space efficiency of the search. Empirical results show that the new techniques significantly improved the efficiency and scalability of A* and BFBnB on most of the benchmark datasets we tested.

The remainder of this paper is structured as follows. Section 2 provides an overview of Bayesian network structure learning and the shortest-path finding formulation of the problem. Section 3 introduces the improved heuristic. Section 4 introduces the sparse representation of optimal parent choices and discusses how to adapt A* and BFBnB algorithms to use the new techniques. Section 5 reports the empirical results on a set of benchmark datasets. Finally, Section 6 concludes the paper with some remarks.

## 2 Background

This section reviews the basics of score-based methods for learning Bayesian network structures.

### 2.1 Learning Bayesian network structures

A Bayesian network is a directed acyclic graph (DAG) in which the vertices correspond to a set of random variables $\mathbf{V} = \{X_1, ..., X_n\}$, and the arcs and lack of them represent dependence and conditional independence relations between the variables. The relations are further quantified using a set of conditional probability distributions. We consider the problem of learning a network structure from a dataset $\mathbf{D} = \{D_1, ..., D_N\}$, where $D_i$ is an instantiation of all the variables in $\mathbf{V}$. A scoring function can be used to measure the goodness of fit of a network structure to $\mathbf{D}$. For example, the minimum description length (MDL) scoring function (Rissanen 1978) uses one term to reward structures with low entropy and another to penalize complex structures. The task is to find an optimal structure that minimizes the MDL score. MDL is decomposable (Heckerman 1998), i.e., the score for a structure is simply the sum of the scores for each variable. All algorithms we describe here assume the scoring function is decomposable. The remainder of the paper assumes the use of MDL score, but our method is equally applicable to other decomposable scoring functions, such as AIC, BIC or BDe.

### 2.2 The shortest-path finding formulation

Yuan *et al.* (2011) formulated the above structure learning problem as a shortest-path finding problem. Figure 1 shows the *implicit* search graph for four variables. The top-most node with the empty set is the *start* search node, and the bottom-most node with the complete set is the *goal* node. An arc from $\mathbf{U}$ to $\mathbf{U} \cup \{X\}$ in the graph represents generating a successor node by adding a new variable $\{X\}$ to the existing variables $\mathbf{U}$; the cost of the arc is equal to the cost of selecting the optimal parent set for $X$ out of $\mathbf{U}$, and is computed by considering all subsets of $\mathbf{U}$, i.e.,

$$BestScore(X, \mathbf{U}) = \min_{PA_X \subseteq \mathbf{U}} score(X|PA_X).$$

With the search graph thus specified, each path from the start node to the goal is an ordering of the variables in the order of their appearance. That is why the search graph is also called an *order graph*. Because each variable only selects optimal parents from the preceding variables, putting together all the optimal parent choices of a particular ordering generates a valid Bayesian network that is optimal for that specific ordering. The shortest path among all possible paths corresponds to a global optimal Bayesian network.

During the search of the order graph, we need to compute the cost for each arc being visited. We use another data structure called *parent graph* for retrieving the costs. The parent graph for variable $X$ consists of all subsets of $\mathbf{V} \setminus \{X\}$. Figure 2 shows the parent graph for $X_1$. Figure 2(a) shows a parent graph containing the raw scores for using each subset as the parent set of $X_1$, while Figure 2(b) shows the optimal scores after propagating the best scores from top to bottom in the graph. For the arc from $\mathbf{U}$ to $\mathbf{U} \cup \{X\}$, we find its score by looking up the parent graph of variable $X$ to find the node that contains $\mathbf{U}$.

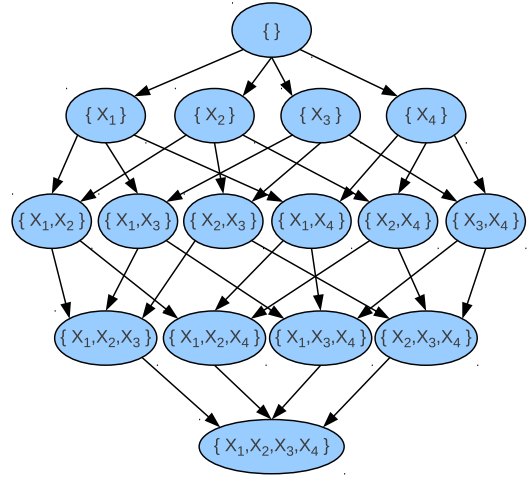

Figure 1: An order graph of four variables

Various search methods as well as dynamic programming have been applied to solve the shortest-path finding problem (Malone, Yuan, and Hansen 2011; Malone et al. 2011; Yuan, Malone, and Wu 2011). In (Yuan, Malone, and Wu 2011), an A* search algorithm was proposed based on the following admissible heuristic function.

**Definition 1.** *Let* $\mathbf{U}$ *be a node in the order graph, its heuristic value is*

$$h(\mathbf{U}) = \sum_{X \in \mathbf{V} \setminus \mathbf{U}} BestScore(X, \mathbf{V} \setminus \{X\}). \quad (1)$$

The A* algorithm is shown to be much more efficient than existing dynamic programming algorithms. However, A* requires all the search information, including parent and order graphs, to be stored in RAM during the search, which makes the algorithm easily run out of memory for large datasets. Malone et al. (2011) developed a breadth-first branch and bound (BFBnB) algorithm to search the order graph in a layered, breadth-first order. By carefully coordinating the parent and order graphs, most of the search information can be stored on disk and are only processed incrementally after being read back to RAM when necessary. The BFBnB algorithm was shown to be as efficient as the A* algorithm but was able to scale to much larger datasets. Theoretically, the scalability of the BFBnB algorithm is only limited by the amount of disk space available.

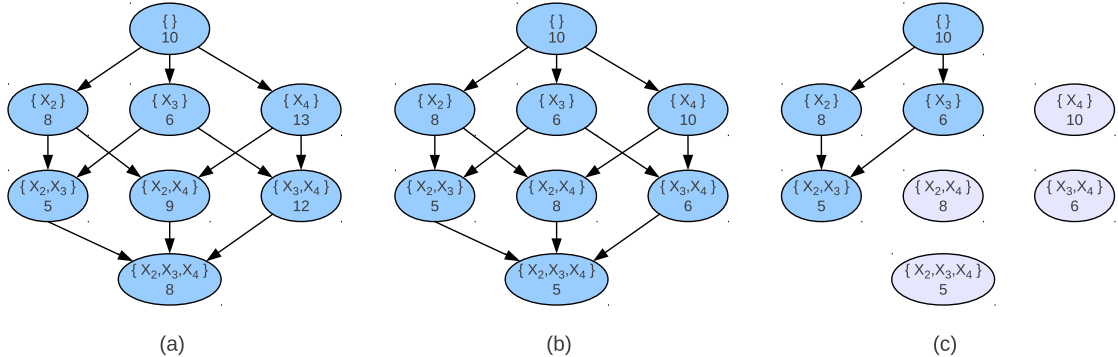

Figure 2: A sample parent graph for variable $X_1$. (a) The raw scores for all the parent sets. The first line in each node gives the parent set, and the second line gives the score of using all of that set as the parents for $X_1$. (b) The optimal scores for each candidate parent set. The second line in each node gives the optimal score using some subset of the variables in the first line as parents for $X_1$. (c) The unique optimal parent sets and their scores. The pruned parent sets are shown in gray. A parent set is pruned if any of its predecessors has an equal or better score.

## 3 An Improved Admissible Heuristic

The heuristic function defined in Equation 1 is based on a classic approach to designing admissible heuristics. Pearl (1984) pointed out that the optimal solution to a relaxed problem can be used as an admissible bound for the original problem. For structure learning, the original problem is to learn a Bayesian network that is an *acyclic* directed graph (DAG). Equation 1 relaxes the problem by completely ignoring the acyclicity constraint, so all *directed graphs* are allowed. This paper aims to improve the heuristic by enforcing partial acyclicity. We will first motivate our approach using a small example. We then describe the specifics of the new heuristic.

### 3.1 A motivating example

According to Equation 1, the heuristic estimate of the start node in the order graph allows each variable to choose optimal parents from all the other variables. Suppose the optimal parents for $X_1, X_2, X_3, X_4$ are $\{X_2, X_3, X_4\}$, $\{X_1, X_4\}$, $\{X_2\}$, $\{X_2, X_3\}$ respectively. These parent choices are shown as the directed graph in Figure 3. Since the acyclicity constraint is ignored, directed cycles are introduced, e.g., between $X_1$ and $X_2$. However, we know the final solution cannot have cycles; three scenarios are possible between $X_1$ and $X_2$: (1) $X_2$ is a parent of $X_1$ (so $X_1$ cannot be a parent of $X_2$), (2) $X_1$ is a parent of $X_2$, or (3) neither of the above is true. The third case is dominated by the other two cases because of the following theorem.

**Theorem 1.** *Let* **U** *and* **V** *be two candidate parent sets for* $X$, *and* $\mathbf{U} \subset \mathbf{V}$, *then* $BestScore(X, \mathbf{V}) \leq BestScore(X, \mathbf{U})$.

This theorem has appeared in many earlier papers, e.g. (Teyssier and Koller 2005; de Campos and Ji 2010), and simply means that an equal or better score can be ob-tained if a larger set of parent candidates is available to choose from. The third case cannot provide a better value than the other two cases because one of the variables must have fewer parents to choose from. Between the first two cases it is unclear which one is better, so we take the minimum of them. Consider the first case first: We have to delete the arc $X_1 \rightarrow X_2$ to rule out $X_1$ as a parent of $X_2$. Then we have to let $X_2$ to rechoose optimal parents from $\{X_3, X_4\}$, that is, we must check all the parent sets not including $X_1$; the deletion of the arc alone cannot produce the new bound. The total bound of $X_1$ and $X_2$ is computed by summing together the original bound of $X_1$ and the new bound of $X_2$. We call this total bound $b_1$. The second case is handled similarly; we call that total bound $b_2$. Because the joint heuristic for $X_1$ and $X_2$ must be optimistic, we compute it as the minimum of $b_1$ and $b_2$. Effectively we have considered all possible ways to break the cycle and obtained a new but improved heuristic value. The new heuristic is clearly admissible, as we still allow cycles among other variables.

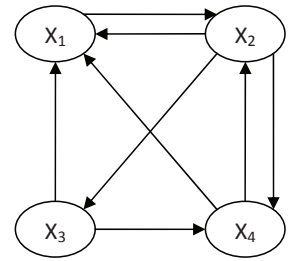

Figure 3: A directed graph representing the heuristic estimate for the start search node.

Often, the simple heuristic introduces multiple cycles. The graph in Figure 3 also has a cycle between $X_2$ and $X_4$. This cycle shares $X_2$ with the earlier cycle; we say they

*overlap*. Overlapping cycles cannot be broken independently. For example, suppose we break the cycle between $X_1$ and $X_2$ by setting the parents of $X_2$ to be $\{X_3\}$. This effectively breaks the cycle between $X_2$ and $X_4$ as well, but introduces a new cycle between $X_2$ and $X_3$. As described in more detail shortly, we divide the variables into non-overlapping groups and focus only on avoiding cycles within each group. So if $X_2$ and $X_3$ are in different groups, they are allowed to form a cycle.

### 3.2 The $k$-cycle conflict heuristic

The idea above can be generalized to compute the joint heuristics for all groups of variables with a size up to $k$ by avoiding cycles within each group. We call the resulting technique the *k-cycle conflict heuristic*. Note that Equation 1 is a special case of this new heuristic, as it simply contains heuristics for the individual variables ($k$=1). The new heuristic is an application of the *additive pattern database* technique (Felner, Korf, and Hanan 2004). We first have to explain what *pattern database* (Culberson and Schaeffer 1998) is. Pattern database is an approach to computing an admissible heuristic for a problem by solving a relaxed problem. As an example, the 15 sliding tile puzzle can be relaxed to only contain the tiles 1-8 by removing the other tiles. Because of the relaxation, multiple states of the original problem are mapped to one state in the *abstract* state space of the relaxed problem. Each abstract state is called a *pattern*, and the exact costs for solving all the abstract states are stored in a pattern database; each cost can be retrieved as an admissible heuristic for any consistent state in the original state space. Furthermore, we can relax the problem in different ways and obtain multiple pattern databases. If the solutions to a set of relaxed problems are independent, the problems are said to have no interactions between them. Again consider 15-puzzle, we can also relax it to only contain the tiles 9-15. This relaxation can be solved independently from the previous one, as they do not share puzzle movements. The costs of their pattern databases can be *added* together to obtain an admissible heuristic, hence the name additive pattern databases.

In the learning problem, a pattern is defined as a set of variables, and its cost is the joint heuristic of the variables involved. The costs of two patterns sharing no variables can be added together to obtain an admissible heuristic because of the decomposability of the scoring function.

We do not need to explicitly break cycles to compute the $k$-cycle conflict heuristic. The following theorem offers a straightforward approach to computing the heuristic.

**Theorem 2.** *The cost of the pattern* $\mathbf{U}$ *is equal to the shortest distance from the node* $\mathbf{V} \setminus \mathbf{U}$ *to the goal node in the order graph.*

The theorem can be proven by noting that avoiding cycles between the variables in $\mathbf{U}$ is equivalent to finding an optimal ordering of the variables with the best joint score, and the different paths from $\mathbf{V} \setminus \mathbf{U}$ to the goal correspond to the different orderings of the variables, among which the shortest path thus corresponds to the optimal ordering. Again consider the example in Figure 3. The joint heuristic for the pattern $\{X_1, X_2\}$ is equal to the shortest distance from the node $\{X_3, X_4\}$ to the goal in Figure 1. Therefore, the new heuristic can be computed by finding the shortest distances from all the nodes in the last $k$ layers of the order graph to the goal. We will describe a backward search algorithm for computing the heuristic in Section 4.2.

Furthermore, the *difference* between the cost of the pattern $\mathbf{U}$ and the simple heuristic of $\mathbf{V} \setminus \mathbf{U}$ indicates the amount of improvement brought by avoiding cycles within the pattern. The differential cost can thus be used as a quality measure for ordering the patterns and for choosing patterns that are more likely to result in a tighter heuristic. Also, we can discard any pattern that does not introduce additional improvement over any of its subset patterns. The pruning can significantly reduce the size of a pattern database and improve the efficiency of accessing the database.

### 3.3 Computing the heuristic for a search node

Once the $k$-cycle conflict heuristic is computed, we can use it to calculate the heuristic value for any node during the search. For a node $\mathbf{U}$, we need to partition $\mathbf{V} \setminus \mathbf{U}$ into a set of non-overlapping patterns, and sum their costs together as the heuristic value. There are potentially many ways to do the partition; ideally we want to find the one with the highest total cost, which represents the most accurate heuristic value. The problem of finding the optimal partition can be formulated as *maximum weighted matching* problem (Felner, Korf, and Hanan 2004). For $k = 2$, we can define a graph in which each vertex represents a variable, and each edge between two variables representing the pattern containing the same variables with an edge weight equal to the cost of the pattern. The goal is to select a set of edges from the graph so that no two edges share a vertex and the total weight of the edges is maximized. The matching problem can be solved in $O(n^3)$ time (Papadimitriou and Steiglitz 1982), where $n$ is the number of vertices.

For $k > 2$, we have to add *hyperedges* to the matching graph for connecting up to $k$ vertices to represent larger patterns. The goal becomes to select a set of edges and hyperedges to maximize the total weight. However, the three-dimensional or higher-order maximum weighted matching problem is NP-hard (Garey and Johnson 1979). That means we have to solve an NP-hard problem when calculating the heuristic value for a search node.

To alleviate the potential inefficiency, we elect to use a greedy method to compute the heuristic value. The method sequentially chooses patterns based on their quality. Consider the node $\mathbf{U}$; the unsearched variables are $\mathbf{V} \setminus \mathbf{U}$. We first choose the pattern with the highest differential cost from all patterns that are subsets of $\mathbf{V} \setminus \mathbf{U}$. We repeat this process by choosing the next pattern for the remaining unsearched variables until all the variables are covered. The total cost of the chosen patterns is used as the heuristic value for the node $\mathbf{U}$.

### 3.4 Dynamic and static pattern databases

The version of the $k$-cycle conflict heuristic introduced above is an example of the *dynamically partitioned pattern database* (Felner, Korf, and Hanan 2004), as the patterns are dynamically selected during the search algorithm. We refer to it as *dynamic pattern database* for short. A potential drawback of dynamic pattern databases is that, even using the greedy method, computing a heuristic values is still more expensive than the simple heuristic in Equation 1. Consequently, the running time can be longer even though the tighter heuristic results in more pruning and fewer expanded nodes.

We can resort to another version of the $k$-cycle conflict heuristic based on the *statically partitioned pattern database* technique (Felner, Korf, and Hanan 2004). The idea is to statically divide all the variables into several groups, and create a separate pattern database for each group. Consider a problem with variables $\{X_1, ..., X_8\}$. We simply divide the variables into two equal-sized groups, $\{X_1, ..., X_4\}$ and $\{X_5, ..., X_8\}$. For each group, say $\{X_1, ..., X_4\}$, we create a pattern database that contains the costs of *all* subsets of $\{X_1, ..., X_4\}$ and store them as a hash table. We refer to this heuristic as the *static pattern database* for short.

It is much simpler to use static pattern databases to compute a heuristic value. Consider the node $\{X_1, X_4, X_8\}$; the unsearched variables are $\{X_2, X_3, X_5, X_6, X_7\}$. We divide these variables into two patterns $\{X_2, X_3\}$ and $\{X_5, X_6, X_7\}$ according to the static grouping. We then simply look up the costs of these two patterns in the pattern databases and sum them together as as the heuristic value for the node. Better yet, every search step only processes one variable and affects one pattern, so computing the heuristic value can be done incrementally.

## 4 The Search Algorithms

Both computing the $k$-cycle conflict heuristic and solving the shortest-path finding problem requires us to search the order graph. The searches further require the parent graphs to be calculated in advance or during the search. In this section, we first introduce a sparse representation of the parent graphs. We then discuss how to search the order graph backward to compute the $k$-cycle conflict heuristic, and forward to solve the shortest path-finding problem by adapting the A* and BFBnB algorithms.

| $parents_{X_1}$ | $\{X_2, X_3\}$ | $\{X_3\}$ | $\{X_2\}$ | $\{\}$ |
|---|---|---|---|---|
| $scores_{X_1}$ | 5 | 6 | 8 | 10 |

Table 1: Sorted scores and parent sets for $X_1$ after pruning parent sets which are not possibly optimal.

| $parents_{X_1}$ | $\{X_2, X_3\}$ | $\{X_3\}$ | $\{X_2\}$ | $\{\}$ |
|---|---|---|---|---|
| $X_2$ | 1 | 0 | 1 | 0 |
| $X_3$ | 1 | 1 | 0 | 0 |
| $X_4$ | 0 | 0 | 0 | 0 |

Table 2: The $parents_X(X_i)$ bit vectors for $X_1$. A "1" in line $X_i$ indicates that the corresponding parent set includes variable $X_i$, while a "0" indicates otherwise. Note that, after pruning, none of the optimal parent sets include $X_4$.

### 4.1 Sparse representation of parent graphs

The parent graph for each variable $X$ exhaustively enumerates the optimal scores for all subsets of $\mathbf{V} \setminus \{X\}$. Naively, this approach requires storing $n2^{n-1}$ scores and parent sets. Due to Theorem 1, however, the number of *unique* optimal parent sets is often far smaller. For example, Figure 2(b) shows that each score may be shared by several nodes in a parent graph. The parent graph representation will allocate space for this repetitive information, resulting in waste of space.

Instead of storing the complete parent graphs, we propose a sparse representation which *sorts* all the *unique* parent scores for each variable $X$ in a list, and also maintain a parallel list that stores the associated optimal parent sets. We call these sorted lists $scores_X$ and $parents_X$. Table 1 shows the sorted lists for the parent graph in Figure 2(b). In essence, this allows us to store and efficiently process only scores in Figure 2(c). We do not have to create the full parent graphs before realizing some scores can be pruned (post-pruning). For example, we can use the following theorem (Tian 2000) to prune some scores before even computing them (pre-pruning).

**Theorem 3.** *In an optimal Bayesian network based on the MDL scoring function, each variable has at most* $\log(\frac{2N}{\log N})$ *parents, where $N$ is the number of data points.*

Because of the pruning of duplicate scores, the sparse representation requires much less memory than storing all the possible parent sets and scores. As long as $\|scores(X)\| < C(n-1, \frac{n}{2})$, it also requires less memory than the BFBnB algorithm for $X$. In practice, $\|scores_X\|$ is almost always smaller than $C(n-1, \frac{n}{2})$ by several orders of magnitude. So this approach offers (usually substantial) memory savings compared to previous best approaches. In addition, the sparse representation is also much more efficient to create because of the pre-pruning.

The key operation in parent graphs is querying the opti-mal parents for variable $X$ out of a candidate set $\mathbf{U}$. With the sparse representation, we can simply scan the list of $X$ starting from the beginning. As soon as we find the first parent set that is a subset of $\mathbf{U}$, we find the optimal parent set and its score. However, scanning the lists can be inefficient if not done properly. Since we have to do the scanning for each arc, the inefficiency will have a large impact on the whole search algorithm. We therefore propose the following incremental approach. Initially, we allow each variable $X$ to use all the other variables as candidate parents, so the first element in the sorted score list must be optimal. For example, the first score in Table 1 must be $BestScore(X_1, \{X_2, X_3, X_4\})$. Suppose we remove $X_2$ from consideration as a candidate parent; we scan the list by continuing from where we last stopped and find a parent set which does not include $X_2$, which must be $BestScore(X_1, \{X_3, X_4\})$ ($\{X_3\}$ in this example). If we further remove $X_3$, we continue scanning the list until finding a parent set which includes neither $X_2$ nor $X_3$ to find $BestScore(X_1, \{X_4\})$ ($\{\}$ it is).

| $parents_{X_1}$ | $\{X_2, X_3\}$ | $\{X_3\}$ | $\{X_2\}$ | $\{\}$ |
|---|---|---|---|---|
| $valid_{X_1}$ | 1 | 1 | 1 | 1 |
| $\sim X_3$ | 0 | 0 | 1 | 1 |
| $valid_{X_1}^{new}$ | 0 | 0 | 1 | 1 |

Table 3: The result of performing the bitwise operation to exclude all parent sets which include $X_3$. A "1" in the $valid_{X_1}$ bit vector means that the parent set does not include $X_3$ and can be used for selecting the optimal parents. The first set bit indicates the best possible score and parent set.

| $parents_{X_1}$ | $\{X_2, X_3\}$ | $\{X_3\}$ | $\{X_2\}$ | $\{\}$ |
|---|---|---|---|---|
| $valid_{X_1}$ | 0 | 0 | 1 | 1 |
| $\sim X_2$ | 0 | 1 | 0 | 1 |
| $valid_{X_1}^{new}$ | 0 | 0 | 0 | 1 |

Table 4: The result of performing the bitwise operation to exclude all parent sets which include either $X_3$ or $X_2$. A "1" in the $valid_{X_1}^{new}$ bit vector means that the parent set includes neither $X_2$ nor $X_3$. The initial $valid_{X_1}$ bit vector had already excluded $X_3$, so finding $valid_{X_1}^{new}$ only required excluding $X_2$.

To further improve the efficiency, we propose the following efficient scanning technique. For each variable $X$, we first initialize an incumbent bit vector of length $\|scores_X\|$ called $valid_X$ to be all 1s. This indicates that all the parent scores in $scores_X$ are usable; the first score in the list will be the optimal score. Then, we create $n-1$ bit vectors also of length $\|scores_X\|$, one for each variable in $\mathbf{V} \setminus \{X\}$. The bit vector for variable $Y$ is denoted as $parents_X(Y)$ and contains 1s for all the parent sets that contain $Y$ and 0s for others. Table 2 shows the bit vectors for Table 1. Then, to exclude variable $Y$ as a candidate par-ent, we perform the bit operation $valid_X^{new} \leftarrow valid_X \& \sim parents_X(Y)$. The $valid_X^{new}$ bit vector now contains 1s for all the parent sets that are subsets of $\mathbf{V} \setminus \{Y\}$. The *first set bit* corresponds to $BestScore(X, \mathbf{V} \setminus \{Y\})$. Table 3 shows an example of excluding $X_3$ from the set of possible parents for $X_1$, and the first set bit in the new bit vector corresponds to $BestScore(X_1, \mathbf{V} \setminus \{X_3\})$. If we further exclude $X_2$, the bit vector resulting from the last step becomes the incumbent bit vector, and a similar bit operation is applied: $valid_X^{new} \leftarrow valid_X \& \sim parents_{X_1}(X_2)$. The first set bit of the result corresponds to $BestScore(X_1, \mathbf{V} \setminus \{X_2, X_3\})$. Table 4 demonstrates this operation. Also, it is important to note that we exclude one variable at a time. For example, if, after excluding $X_3$, we wanted to exclude $X_4$ rather than $X_2$, we could take $valid_X^{new} \leftarrow valid_X \& \sim parents_X(X_4)$.

### 4.2 Creating the $k$-cycle conflict heuristic

We have two versions of the $k$-cycle conflict heuristic: dynamic and static pattern databases. To compute the dynamic pattern database, we use the *breadth-first search* to do a backward search for $k$ layers in the order graph. The search starts from the goal node and expands the order graph backward layer by layer. A reverse arc from $\mathbf{U} \cup \{X\}$ to $\mathbf{U}$ has a cost equal to $BestScore(X, \mathbf{U})$. The reverse $g$ cost to $\mathbf{U}$ is updated whenever a new path with a lower cost is found. Breadth-first search ensures that node $\mathbf{U}$ will obtain its optimal reverse $g$ cost once the whole layer is processed. Its corresponding pattern $\mathbf{V} \setminus \mathbf{U}$ is pruned if the differential score is equal to that of any subset pattern. Otherwise, it is added to the pattern database together with both its pattern cost and differential cost.

The static pattern databases are calculated differently. For a static grouping $\mathbf{V} = \bigcup_i \mathbf{V_i}$, we need to compute a pattern database for each group $\mathbf{V_i}$, which is basically a full order graph containing all subsets of $\mathbf{V_i}$. We will also use a backward breadth first search to create the graph layer by layer starting from the node $\mathbf{V_i}$. However, the cost for any reverse arc from $\mathbf{U} \cup \{X\}$ to $\mathbf{U}$ in this order graph will be $BestScore(X, (\bigcup_{j \neq i} \mathbf{V_j}) \cup \mathbf{U})$.

### 4.3 Solving the shortest-path finding problem

After the pattern database heuristics are computed, we solve the shortest-path finding problem using a forward search in the order graph. We adapt both the A* (Yuan, Malone, and Wu 2011) and BFBnB (Malone, Yuan, and Hansen 2011) algorithms to utilize the new heuristic and the sparse parent graphs.

Originally, the A* algorithm first creates the full parent graphs and then expands the order graph in a best-first order starting from the top. For the improved version, we first create the unique score lists and the $k$-cycle conflict heuristic. During the search, the only difference appears in gener-ating the successors of a node. For each successor $\mathbf{U} \cup \{X\}$ of node $\mathbf{U}$, we calculate its heuristic value according to the methods described in Sections 3.3 and 3.4. Looking up the cost $BestScore(X, \mathbf{U})$ for the arc $\mathbf{U} \rightarrow \mathbf{U} \cup \{X\}$ is achieved by using the sparse parent graphs.

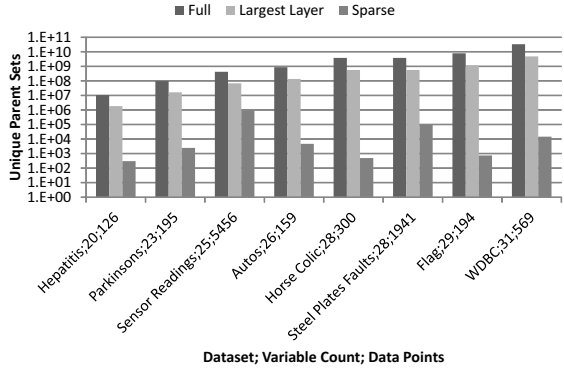

Figure 4: The number of parent sets stored in the full parent graphs ("Full"), the largest layer of the parent graphs ("Largest Layer"), and the sparse representation ("Sparse").

The BFBnB algorithm is affected in a similar way. Originally it works by coordinating the expansion of the order graph and parent graphs layer by layer. In the improved version, the unique score lists and the heuristic are calculated first. The search part of the algorithm only needs to expand the order graph, during which generating successors works similarly as in the improved A* algorithm.

## 5 Empirical Results

We tested our new techniques on the A* and BFBnB algorithms by comparing to their original versions[1]. The experiments were performed on a PC with 3.07 GHz Intel i7 processor, 16 GB of RAM, 500 GB of hard disk space, and running Ubuntu 10.10. We used benchmark datasets from the UCI machine learning repository (Frank and Asuncion 2010) to test the algorithms. For all the datasets, records with missing values were removed. All variables were discretized into two states around their means.

### 5.1 Memory savings of sparse parent graphs

We first evaluated the memory savings made possible by using the sparse representation in comparison to the full parent graphs. In particular, we compared the maximum number of scores that have to be stored for all variables at once by each algorithm. A typical dynamic programming algorithm stores scores for all possible parent sets of all variables. BFBnB and memory-efficient dynamic programming (Malone, Yuan, and Hansen 2011) (assuming implementation optimizations) store all possible parent sets only in one layer of the parent graphs for all variables, so the size of the largest layer of all parent graphs is an indication of its space requirement. The sparse representation only stores the unique optimal parent sets for all variables at all layers.

Figure 4 shows the memory savings by the sparse representation. The number of unique scores stored by the sparse representation is typically several orders of magnitude smaller than the number of parent sets stored by the full representation. These results agree quite well with previously published results (de Campos and Ji 2010).

Due to Theorem 3, increasing the number of data records increases the maximum number of candidate parents. Therefore, the number of unique candidate parent sets increases as the number of records increases; however, many of the new parent sets are pruned. The number of variables also affects the number of candidate parent sets. Consequently, the number of unique scores increases as a function of the number of records and the number of variables. The amount of pruning is data-dependent, though, and is not easily predictable. In practice, we find the number of records to affect the number of unique scores more than the number of variables. Other scoring functions, such as BDe, exhibit similar behavior.

The results also suggest that the savings increase as the number of variables increases in the datasets. This implies that, while more variables necessarily increases the number of possible parent sets exponentially, the number of unique optimal parent sets increases much more slowly. Intuitively, even though we add more parents, only a small number of them are "good" parents for any particular variable.

### 5.2 Heuristics vs sparse representation

Both the new heuristic and the sparse representation can be used to improve the A* and BFBnB algorithms. It is beneficial to have an understanding on how much improvement each technique contributes. Also, the new heuristic has two versions: static and dynamic pattern databases; each of them can be parameterized in different ways. We applied various parameterizations of the new techniques to the algorithms on the datasets Autos and Flag. For the dynamic pattern database, we varied $k$ from 2 to 4. For the static pattern databases, we tried groupings 9-9-8 and 13-13 for the Autos dataset and groupings 10-10-9 and 15-14 for the Flag dataset. The results are shown in Table 5.

The sparse representation helped both A* and BFBnB algorithms to achieve much better efficiency and scalability. A* ran out of memory on both of the datasets when using full parent graphs, but was able to solve both Autos (with any heuristic) and Flag (with some of the best heuristics) when using sparse parent graphs. Similarly, BFBnB ran out of time on the Flag dataset within the one hour time limit when using full parent graphs, but was able to solve the dataset using the sparse representation (except when using the simple heuristic); on Autos, the sparse representation helped improve the time efficiency of BFBnB by up to an order of magnitude. One last note here is the numbers of expanded nodes by BFBnB are slightly different when using the two representations; it is only because of the randomness in the local search method used to compute the initial upper bound solution for BFBnB.

---
[1] A software package named *URLearning* ("you are learning") implementing the A* and BFBnB algorithms can be downloaded at http://url.cs.qc.cuny.edu/software.html.

|         | Pattern Database |        | BFBnB, Full |            | BFBnB, Sparse |             | A*, Sparse |            |
| Dataset | Type             | Size   | Time (s)    | Nodes      | Time (s)      | Nodes       | Time (s)   | Nodes      |
|---------|------------------|--------|-------------|------------|---------------|-------------|------------|------------|
| Autos   | Simple           | 26     | 2,690       | 62,721,601 | 461           | 62,721,601  | 674        | 35,329,016 |
| Autos   | Dynamic, k=2     | 41     | 2,722       | 52,719,774 | 449           | 52,719,793  | 148        | 6,286,142  |
| Autos   | Dynamic, k=3     | 116    | 2,720       | 49,271,793 | 468           | 49,271,809  | 76         | 2,829,877  |
| Autos   | Dynamic, k=4     | 582    | 2,926       | 48,057,187 | 699           | 48,057,205  | 67         | 2,160,515  |
| Autos   | Static, 9-9-8    | 1,280  | 2,782       | 57,002,704 | 495           | 57,002,715  | 228        | 9,763,518  |
| Autos   | Static, 13-13    | 16,384 | 2,747       | 48,814,324 | 211           | 48,814,334  | 125        | 4,762,276  |
| Flag    | Simple           | 29     | OT          | OT         | OT            | OT          | OM         | OM         |
| Flag    | Dynamic, k=2     | 45     | OT          | OT         | 1,222         | 132,431,610 | 824        | 19,359,296 |
| Flag    | Dynamic, k=3     | 149    | OT          | OT         | 788           | 79,332,390  | 207        | 5,355,085  |
| Flag    | Dynamic, k=4     | 858    | OT          | OT         | 1,624         | 84,054,443  | 350        | 7,377,817  |
| Flag    | Static, 10-10-9  | 2,560  | OT          | OT         | 2,600         | 249,638,318 | OM         | OM         |
| Flag    | Static, 15-14    | 49,152 | OT          | OT         | 720           | 88,305,173  | 136        | 4,412,232  |

Table 5: A comparison of the enhanced A* and BFBnB algorithms with various combinations of parent graph representations (full vs. sparse) and the heuristics (simple heuristic, dynamic pattern database with $k = 2, 3$, and 4, and static pattern databases with groupings 9-9-8 and 13-13 for the Autos dataset and groupings 10-10-9 and 15-14 for the Flag dataset). "Size" means the number of patterns stored; "Sparse" means the sparse parent graphs; "Full" means the full parent graphs; "Time" means the running time (in seconds), and "Nodes" means the number of nodes expanded by the algorithms; "OT" means the algorithm fail to finish within a 1-hour time limit set for this experiment; and "OM" means the algorithm used up all the RAM (16G). "A*, Full" is not included because it ran out of memory in all cases.

Both the static and dynamic pattern databases helped A* and BFBnB algorithms to improve efficiency and scalability. A* with both the simple heuristic and the static pattern database with grouping 10-10-9 ran out of memory on the Flag dataset. The other pattern database heuristics enabled A* to finish successfully. The dynamic pattern database with $k = 2$ helped to reduce the number of expanded nodes significantly for both algorithms on the datasets. Setting $k = 3$ helped even more. However, further increasing $k$ to 4 often resulted in increased running time, and sometimes an increased number of expanded nodes as well. We believe that a larger $k$ always results in a better heuristic; the occasional increase in expanded nodes is because the greedy strategy we used to choose patterns did not fully utilize the larger pattern database. The longer running time is reasonable though because it is less efficient to compute a heuristic value in larger pattern databases, and the inefficiency gradually overtook the benefit brought by the better heuristic. Therefore, $k = 3$ seems to be the best parametrization for the dynamic pattern database in general. For the static pattern databases, we were able to test much larger groups as we do not need to enumerate all groups up to a certain size. The results suggest that larger groupings tend to result in tighter heuristic.

The sizes of the static pattern databases are typically much larger than the dynamic pattern databases. However, they are still negligible in comparison to the number of expanded search nodes in all cases. It is thus cost effective to try to compute larger but affordable-size static pattern databases to achieve better search efficiency. The results show that the best static pattern databases typically helped A* and BFBnB to achieve better efficiency than the best dynamic pattern database, even when the number of expanded nodes is larger. The reason is calculating the heuristic values is more efficient when using static pattern databases.

### 5.3 Results on other datasets

Since static pattern databases seem to work better than dynamic pattern databases in most cases, we tested A* and BFBnB using static pattern database and sparse representation on all the datasets against the original algorithms. We used the simple static grouping of $\lceil \frac{n}{2} \rceil - \lfloor \frac{n}{2} \rfloor$ for all the datasets, where $n$ is the number of variables. The results are shown in Table 6.

For the BFBnB algorithm, the improved version was around 5 times faster than the original version and sometimes even orders of magnitude faster (e.g. Flag). The reduction in the number of expanded nodes is not as dra-matic, however. The main reason is that the original BFBnB algorithm interleaves expanding the order graph and the full parent graphs during the search, while the improved version first calculates the sparse representation of parent scores, and then performs the search. It is much more efficient to compute the sparse representation than computing the full parent graphs. However, on the dataset Sensor Readings, the improved BFBnB algorithm runs slower than the original version. There are two potential explanations. First, this particular dataset has a large number of data points, which makes the sparse representation not truly sparse. Second, the new heuristic seems to be not much tighter than the simple heuristic on this dataset, because the numbers of expanded nodes are very similar in both cases.

| Dataset | | | Results | | | | |
|---|---|---|---|---|---|---|---|
| Name | n | N | | BFBnB | BFBnB (SP) | A* | A* (SP) |
| Hepatitis | 20 | 126 | Time (s) | 9 | 1 | 6 | 0 |
| | | | Nodes | 610,974 | 129,889 | 411,150 | 8,565 |
| Parkinsons | 23 | 195 | Time (s) | 100 | 19 | 100 | 15 |
| | | | Nodes | 8,388,607 | 4,646,877 | 8,388,607 | 1,152,576 |
| Sensor Readings | 25 | 5,456 | Time (s) | 632 | 3,121 | OM | 731 |
| | | | Nodes | 33,554,431 | 33,554,430 | OM | 3,286,650 |
| Autos | 26 | 159 | Time (s) | 1,170 | 211 | OM | 111 |
| | | | Nodes | 53,236,395 | 48,814,295 | OM | 4,762,276 |
| Horse Colic | 28 | 300 | Time (s) | 4,221 | 678 | OM | OM |
| | | | Nodes | 268,435,455 | 74,204,000 | OM | OM |
| Steel Plates Faults | 28 | 1,941 | Time (s) | 7,913 | 4,544 | OM | OM |
| | | | Nodes | 268,435,455 | 264,887,347 | OM | OM |
| Flag | 29 | 194 | Time (s) | 12,902 | 421 | OM | 147 |
| | | | Nodes | 354,388,170 | 88,305,173 | OM | 4,412,232 |
| WDBC | 31 | 569 | Time (s) | 93,382 | 26,196 | OM | OM |
| | | | Nodes | 1,353,762,809 | 273,746,036 | OM | OM |

Table 6: A comparison on the number of nodes expanded and running time (in seconds) of the A* and BFBnB algorithms enhanced by both static pattern database with grouping $\lceil \frac{n}{2} \rceil - \lfloor \frac{n}{2} \rfloor$, where $n$ is the number of variables, and sparse representation of parent scores (denoted by "SP") against the original versions of these algorithms. "n" is the total number of variables, and "N" is the number of data points.

The benefits of the new techniques are more obvious when applied to A*. For the datasets on which the original algorithm was able to finish, the improved algorithm was up to one order of magnitude faster; the number of expanded nodes is also significantly reduced. In addition, it was able to solve three larger datasets: Sensor Readings, Autos, and Flag. The running time on each of those datasets is pretty short, which indicates that once the memory consumption of the parent graphs was reduced, the A* algorithm was able to use more memory for the order graph and solved the search problems rather easily.

## 6 Concluding Remarks

The shortest-path finding formulation of the learning problem presented in (Yuan, Malone, and Wu 2011) makes two orthogonal directions of research natural. One is the development of search algorithms for learning optimal Bayesian networks, represented by the A* and BFBnB algorithms developed in (Yuan, Malone, and Wu 2011; Malone et al. 2011). One contribution of this paper is the sparse representation of the parent graphs which only store the unique optimal parent sets and scores. The method improves the time and space efficiency of the parent graph part of the search and thus falls in the first direction.

The second direction, which we believe is equally important, is the development of search heuristics. Another contribution of this paper is a new admissible heuristic called the $k$-cycle conflict heuristic developed based on the additive pattern databases. We tested the A* and BFBnB algorithms enhanced by the new heuristic and the sparse representation on a set of UCI machine learning datasets. The results show that both of the new techniques contributed to significant improvement in the efficiency and scalability of the algorithms. We therefore believe the new methods represent another significant step forward in exact Bayesian network structure learning.

As future work, we plan to investigate better approaches to obtaining the groupings for the static pattern databases. It could be based on prior knowledge, or some initial estimation of the correlation between the variables. Such groupings are expected to work better than the simple grouping we tested in this paper.

**Acknowledgements** This work was supported by NSF grants IIS-0953723 and EPS-0903787.